\documentclass[submission,copyright,creativecommons]{eptcs}

\providecommand{\event}{QPL 2014} 

\usepackage{latexsym}
\usepackage{lipsum}

\usepackage{amsmath}
\usepackage{amsfonts}
\usepackage{amssymb} 
\usepackage{amsthm}

\usepackage[svgnames]{xcolor}
\usepackage{tikz}
\usepackage{wrapfig}

\pgfdeclarelayer{edgelayer}
\pgfdeclarelayer{nodelayer}
\pgfsetlayers{edgelayer,nodelayer,main}

\tikzstyle{none}=[inner sep=0pt]
\tikzstyle{plain}=[inner sep=0pt]
\tikzstyle{every picture}=[baseline=(current bounding box).east,scale=0.3,node distance=5mm]

\newcommand{\ov}{\overrightarrow} 

\title{A Study of Entanglement in\\a Categorical Framework of Natural Language}

\author{Dimitri Kartsaklis
\institute{University of Oxford\\
Department of Computer Science\\
Oxford, UK}
\email{dimitri.kartsaklis@cs.ox.ac.uk}
\and
Mehrnoosh Sadrzadeh
\institute{Queen Mary University of London\\
School of Electronic Engineering and Computer Science\\
London, UK}
\email{mehrnoosh.sadrzadeh@qmul.ac.uk}
}

\def\titlerunning{A Study of Entanglement in a Categorical Framework of Natural Language}
\def\authorrunning{D. Kartsaklis, M. Sadrzadeh}

\newcommand{\ctikzfig}[1]{%
\begin{center}\rm
  
\InputIfFileExists{#1.tikz}{}{\input{.//#1.tikz}}

\end{center}}

\begin{document}

\maketitle

\begin{abstract}
In both quantum mechanics and corpus linguistics based on vector spaces, the notion of entanglement provides a means for the various subsystems to communicate with each other. In this paper we examine a number of implementations of the categorical framework of Coecke et al. \cite{coeckeetal} for natural language, from an entanglement perspective. Specifically, our goal is to better understand in what way the level of entanglement of the relational tensors (or the lack of it) affects the compositional structures in practical situations. Our findings reveal that a number of proposals for verb construction lead to almost separable tensors, a fact that considerably simplifies the interactions between the words. We examine the ramifications of this fact, and we show that the use of Frobenius algebras mitigates the potential problems to a great extent. Finally, we briefly examine a machine learning method that creates verb tensors exhibiting a sufficient level of entanglement.
\end{abstract}

\section{Introduction}

Category theory  in general and compact closed categories in particular provide a high level framework to identify  and study universal properties of mathematical and physical structures. Abramsky and Coecke \cite{abramsky2004}, for example, use the latter to provide a structural  proof for a class of quantum protocols, essentially recasting the vector space semantics of quantum mechanics in a more abstract way. This and similar kinds of abstraction have made compact closed categories applicable to other fields with vector space semantics, for the case of this paper, corpus linguistics.  Here, Coecke et al.\cite{coeckeetal} used  them to unify two seemingly orthogonal semantic models of natural language: a syntax-driven compositional approach as expressed by Lambek \cite{Lambek} and distributional models of meaning based on vector spaces. The latter approach is capable of providing a concrete representation of the meaning of a word, by creating a vector with co-occurrence counts of that word in a corpus of text with all other words in the vocabulary. Distributional models of this form have been proved useful in many natural language processing tasks \cite{Schutze,Manning,Landauer}, but in general they do not scale up to larger text constituents such as phrases and sentences. On the other hand, the type-logical approaches to language as introduced in \cite{Lambek} are compositional but unable to provide a convincing model of word meaning. 

The unification of the two semantics paradigms is based on the fact that both a type logic expressed as a pregroup \cite{Lambek} and finite dimensional vector spaces share a compact closed structure; so in principle there exists a way to express a grammatical derivation as a morphism that defines mathematical manipulations between vector spaces, resulting in a sentence vector. In \cite{coeckeetal}, the solution was based on a Cartesian product between the pregroup category and the category of finite dimensional vector spaces; later  this was recasted in a functorial passage from the former to the latter \cite{Preller2010,Sadrzadeh2013,Kartsaklis2014ta}. The general idea behind any of these frameworks is that the grammatical type of each word determines the vector space where the corresponding vector lives. Words with atomic types, such as nouns, are simple vectors living in $N$. On the other hand, words with relational types, such as adjectives or verbs, live in tensor product spaces of higher order. For instance, an intransitive verb will be an element of an order-2 space such as $N\otimes S$, whereas a transitive verb will live in $N \otimes S \otimes N$. These tensors act on their arguments by \textit{tensor contraction}, a generalization of the familiar notion of matrix multiplication to higher order tensors.

Since every relational word is represented by a tensor, naturally \textit{entanglement} becomes an important issue in these models. Informally speaking, elements of tensor spaces which represent meanings of relational words should be entangled to allow for a so called `flow of information' (a terminology borrowed from categorical quantum mechanics \cite{abramsky2004}) among the meanings of words in a phrase or sentence. Otherwise, parts of the meaning of these words become isolated from the rest, leading to unwanted consequences. An example would be that all sentences that have the same verb end up to get the same meaning regardless of the rest of the context, and this is obviously not the case in language. Whereas at least intuitively the above argument makes sense, in some of the language tasks we have been experimenting with, non-entangled  tensors have provided very good results. For example, in \cite{Grefenstette2011b} Grefenstette and Sadrzadeh provide results for verbs that are built from the outer product of their context vectors. These results beat the state of the art of that time (obtained by the same authors in a previous paper \cite{GrefenSadr1}) by a considerable difference.  

The purpose of the current paper is to provide a preliminary study of the entanglement in corpus linguistics and to offer some explanation why phenomena such as the above have been the case: is this a by-product of the task or the corpus or the specific concrete model?  We work with a number of concrete instantiations of the framework in sentence similarity tasks  and observe their performances experimentally from an entanglement point of view. Specifically, we investigate a number of models based on the weighted relations method of \cite{GrefenSadr1}, where a verb matrix is computed as the structural mixing of all subject/object pairs with which it appears in the training corpus. We also test a model trained using linear regression \cite{Baroni}. Our findings for the first case have been surprising. It turns out that, contrary to intuition and despite the fact that the construction method should yield entangled matrices, the results are very close to their rank-1 approximations, that is, they are in effect separable. We further investigate the ramifications of this observation and  try to explain the good practical predictions. We then experiment with the linear regression model of \cite{Baroni} and show  that the level of entanglement is much higher in the verbs of this model. Finally, we look at a number of Frobenius variations of the weighted relation models, such as the ones presented in \cite{kartsaklis2012} and a few new constructions exclusive to this paper. The conclusions here are also surprising, but in a positive way. It seems that Frobenius models are able to overcome the unwanted ``no-flow'' collapses of the separable verbs by generating a partial flow between the verb and either its subject or its object, depending which dimension they are copying.  


\section{Quantizing the grammar}

The purpose of the categorical framework is to map a grammatical derivation to some appropriate manipulation between vector spaces. In this section we will shortly review how this goal is achieved. Our basic type logic is a \textit{pregroup grammar} \cite{Lambek}, built on the  basis of a pregroup algebra. This  is a partially ordered monoid with unit 1, whose each element $p$ has a left adjoint $p^l$ and a right adjoint $p^r$. This means that they satisfy the following inequalities:

\begin{equation}
  p^l \cdot p \leq 1~~~~p \cdot p^r \leq 1~~~~\text{and}~~~~1 \leq p \cdot p^l~~~~1 \leq p^r \cdot p
  \label{equ:pregroups}
\end{equation}

A pregroup grammar is the pregroup freely generated over a set of atomic types, which for this paper will be $\{n,s\}$. Here, type $n$ refers to nouns and noun phrases, and type $s$ to sentences. The atomic types and their adjoints can be combined to create types for \emph{relational words}. The type of an adjective, for example, is $n \cdot n^l$, representing something that inputs a noun (from the right) and outputs another noun. Similarly, the type of a transitive verb $n^r \cdot s \cdot n^l$ reflects the fact that verbs of this kind expect two inputs, one noun at each side. A grammatical reduction then follows from the properties of pregroups and specifically the inequalities in (\ref{equ:pregroups}) above. The derivation for the sentence `Happy kids play games' has the following form:

\begin{equation*}
  (n \cdot n^l) \cdot n \cdot (n^r \cdot s \cdot n^l) \cdot n = n \cdot (n^l \cdot n) \cdot n^r \cdot s \cdot (n^l \cdot n)
  \leq n \cdot 1 \cdot n^r \cdot s \cdot 1 = n \cdot n^r \cdot s \leq 1 \cdot s = s
\end{equation*}

We refer to the free pregroup generated by a partially ordered set $T$ as \textbf{Preg$_F$}$(T)$. Categorically, this structure conforms to the definition of a non-symmetric \textit{compact closed category}. The inequalities in (\ref{equ:pregroups}) correspond to the $\epsilon$ and $\eta$ morphisms of a compact closed category, given as follows:

\begin{equation}
\label{equ:epsilon}
  \epsilon^l: A^l \otimes A \to I~~~~~~~~~
  \epsilon^r: A \otimes A^r \to I
\end{equation}

\vspace{-0.5cm}  
\begin{equation}
  \label{equ:eta1}
  \eta^l: I \to A \otimes A^l~~~~~~~~~
  \eta^r:I \to A^r \otimes A
\end{equation}
\vspace{-0.5cm}

\noindent
Hence the above grammatical reduction becomes the  following morphism:

\vspace{-0.2cm}

\begin{equation}
  (\epsilon^r_n \otimes 1_s) \circ (1_n \otimes \epsilon^l_n \otimes 1_{n^r \cdot s} \otimes \epsilon^l_n)
  \label{equ:example}
\end{equation}

Category \textbf{Preg$_F$}$(T)$ is posetal, which means that there is at most one morphism between two given objects. To make this into a full-blown category we work with the free compact closed category generated over $T$, as described in \cite{preller07}, which we will denote $\textbf{C$_F$}(T)$. Furthermore, let us refer to the category of finite-dimensional vector spaces and linear maps over $\mathbb{R}$ as \textbf{FVect$_W$}, where $W$ is our basic distributional vector space with an orthonormal basis $\{w_i\}_i$. This category is again compact closed (although a symmetric one, since $W \cong W^*$), with the $\epsilon$ and $\eta$ maps given as follows:


\begin{equation}
\label{equ:innerprod}
  \epsilon^l = \epsilon^r: W \otimes W \to \mathbb{R}:: \sum\limits_{ij} c_{ij}(\ov{w_i} \otimes \ov{w_j}) \mapsto \sum\limits_{ij} c_{ij}\langle \ov{w_i}|\ov{w_j}\rangle
\end{equation}

\vspace{-0.2cm}
\begin{equation}
\label{equ:eta2}
  \eta^l = \eta^r: \mathbb{R} \to W \otimes W :: 1 \mapsto \sum\limits_i \ov{w_i} \otimes \ov{w_i}
\end{equation}

The transition from a pregroup reduction to a morphism  between vector spaces is achieved by a \textit{strongly monoidal functor} $\mathcal{F}: \textbf{C$_F$}(T) \to \textbf{FVect}_W$ that preserves the compact structure so that
$\mathcal{F}(A^l) = \mathcal{F}(A)^l$ and $\mathcal{F}(A^r) = \mathcal{F}(A)^r$. Further, since $\textbf{FVect}_W$ is symmetric and  $W$ has a fixed basis, we have that $\mathcal{F}(A)^r = \mathcal{F}(A)^l \cong \mathcal{F}(A)$. As motivated in previous work \cite{kartsaklis2012}, we assume that $\mathcal{F}$ assigns the basic vector space $W$ to both of the atomic types, that is we have:

\begin{equation}
   \mathcal{F}(n) = \mathcal{F}(s) = W
\end{equation}

The partial orders between the atomic types are mapped to linear maps from $W$ to $W$ by functoriality. The adjoints of atomic types are also mapped to $W$, whereas the complex types are mapped to tensor products of vector spaces:

\begin{equation}
  \mathcal{F}(n \cdot n^l) = \mathcal{F}(n^r \cdot s) = W \otimes W~~~~~~\mathcal{F}(n^r \cdot s \cdot n^l) = W \otimes W \otimes W
\end{equation}

We are now in position to define the meaning of a sentence $w_1w_2\hdots w_n$ with type reduction $\alpha$ as follows:

\vspace{-0.3cm}
\begin{equation}
  \mathcal{F}(\alpha)(\ov{w_1} \otimes \ov{w_2} \otimes \hdots \otimes \ov{w_n})
\end{equation}

For example,  the meaning of the sentence  `happy kids play games', which has the grammatical reduction  (\ref{equ:example}),  is computed as follows:

\begin{eqnarray*}
  \mathcal{F} \left[ (\epsilon^r_n \otimes 1_s) \circ (1_n \otimes \epsilon^l_n \otimes 1_{n^r \cdot s} \otimes \epsilon^l_n) \right]
 \left( \overline{happy} \otimes \ov{kids} \otimes \overline{play} \otimes \ov{games}  \right) = \\
  (\epsilon_W \otimes 1_W) \circ (1_W \otimes \epsilon_W \otimes 1_{W \otimes W} \otimes \epsilon_W)
 \left( \overline{happy} \otimes \ov{kids} \otimes \overline{play} \otimes \ov{games}  \right)~~~~
\end{eqnarray*}

\noindent
The above categorical computations simplify to the following form:
\begin{equation}
  (\overline{happy} \times \ov{kids})^\mathsf{T} \times \overline{play} \times \ov{games}
  \label{equ:result}
\end{equation}

\noindent
where symbol $\times$ denotes tensor contraction and the above is a vector living in our basic vector space $W$.

\section{Pictorial calculus}

%

Compact closed categories are complete with regard to a pictorial calculus \cite{kelly1972many,selinger2011survey}, which can be used for visualizing the derivations and simplifying the computations. We introduce the fragment of calculus that is relevant to the current paper. A morphism $f:A \to B$ is depicted as a box with incoming and outgoing wires representing the objects; the identity morphism $1_A: A \to A$ is a straight line.

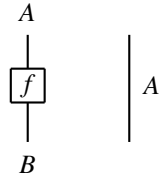
\small
\ctikzfig{morphism1}
\normalsize

%

Recall that the objects of \textbf{FVect$_W$} are vector spaces. However, for our purposes it is also important to access individual vectors within a vector space. In order to do that, we represent a vector $\ov{v} \in V$ as a morphism $\ov{v}: I \to V$. The unit object is depicted as a triangle, while the number of wires emanating from it denotes the order of the corresponding tensor.

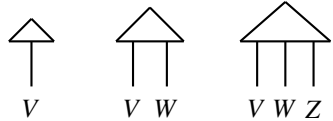
\small
\ctikzfig{samples}
\normalsize

%

Tensor products of objects and morphisms are depicted by juxtaposing the corresponding diagrams side by side. Composition, on the other hand, is represented as a vertical superposition.  For example, from left to right,  here are the pictorial representations of the tensor of  a vector in $A$ with a vector in $B$, a tensor of morphisms $f \otimes g:A \otimes C \to B \otimes D$, and a composition of morphisms $h \circ f$ for $f: A \to B$ and $h: B \to C$:

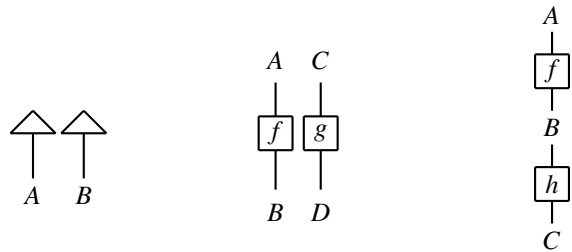
\small
\ctikzfig{tensor}
\normalsize

The $\epsilon$-maps are represented as cups ($\cup$) and the $\eta$-maps as caps ($\cap$). Equations such as $(\epsilon^l_A \otimes 1_{A^r}) \circ (1_{A^l} \otimes \eta^r_A) = 1_A$ now get an intuitive visual justification:

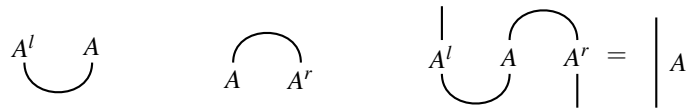
\small
\ctikzfig{morphisms}
\normalsize

%

We are now in position to provide a diagram for the meaning of the sentence `happy kids play games'.

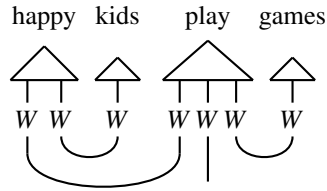
\small
\ctikzfig{sentence}
\normalsize

We conclude this section with one more addition to our calculus. As in most quantum protocols, some times the flow of information in linguistics requires elements of classical processing; specifically, we will want the ability to \textit{copy} and \textit{delete} information, which can be provided by introducing \textit{Frobenius algebras}. In \textbf{FVect}, any vector space $V$ with a fixed basis $\{\ov{v_i}\}$ has a Frobenius algebra over it given by Eqs. \ref{equ:frob} below. 

\vspace{0.3cm}
\begin{minipage}{0.2\linewidth}
\small
\ctikzfig{frob}
\normalsize
\end{minipage}
\begin{minipage}{0.75\linewidth}
\vspace{-0.5cm}
\begin{align}
\label{equ:frob}
  \Delta::\ov{v_i} \mapsto \ov{v_i} \otimes \ov{v_i} & & \iota::\ov{v_i} \mapsto 1 \\ \nonumber
 \mu:: \ov{v_i} \otimes \ov{v_j} \mapsto \delta_{ij}\ov{v_i} :=  \left\{\begin{array}{c l}
   \ov{v_i} & i=j \\ \nonumber
   \ov{0} & i \neq j
\end{array} \right. & &
  \zeta::1 \mapsto \sum_i \ov{v_i}
\end{align}
\end{minipage}

\section{Entanglement in quantum mechanics and linguistics}

Given two non-interacting quantum systems $A$ and $B$, where $A$ is in state $|\psi\rangle_A$ and $B$ in state $|\psi\rangle_B$, we denote  the state of the composite system $A \otimes B$ by $|\psi\rangle_A \otimes |\psi\rangle_B$. States of this form that can be expressed as the tensor product of two state vectors are called \textit{product} states, and they constitute a special case of separable states. In general, however, the state of a composite system is not necessarily a product state or even a separable one. Fixing bases $\{|i\rangle_A\}$ and $\{|j\rangle_B\}$ for the vector spaces of the two states, a general composite  state (separable or not) is denoted as follows:

\begin{equation}
   |\psi\rangle_{AB} = \sum_{ij}c_{ij}|i\rangle_A \otimes |j\rangle_B
\end{equation}

%

\noindent
In the case of a pure quantum state, $|\psi\rangle_{AB}$ is separable only if it can be expressed as the tensor product of two vectors; otherwise it is \textit{entangled}. In a similar way, the tensor of a relational word is separable  if it is equal to the  tensor product of two vectors. In our graphical calculus, these objects  are depicted  by  the juxtaposition of two or more triangles:

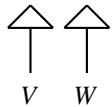
\small
\ctikzfig{separ}
\normalsize

%

In general, a tensor is not separable if it is a linear combination of many separable tensors. The number of separable tensors needed to express the original tensor is equal to  the \textit{tensor rank}. Graphically, a tensor of this form is shown as a single triangle with two or more legs:

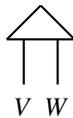
\small
\ctikzfig{entang}
\normalsize

\section{Consequences of separability}
\label{sec:cons}

In categorical quantum mechanics terms, entangled states are necessary to allow the flow of information between the different subsystems. In this section we show that the same is true for linguistics. Consider the diagram of our example derivation, where all relational words are now represented by separable tensors (in other words, no entanglement is present):

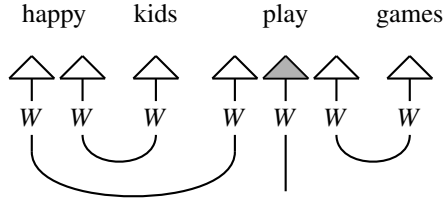
\small
\ctikzfig{sentencesep}
\normalsize

\noindent 
In this version, the  $\epsilon$-maps are  completely detached from the components of the relational tensors that carry the results (left-hand wire of the adjective and middle wire of the verb); as a consequence, flow of information is obstructed, all compositional interactions have been eliminated, and the meaning of the sentence is reduced to the middle component of the verb (shaded vector) multiplied by a scalar, as follows (superscripts denote the left-hand, middle, and right-hand components of separable tensors):

\begin{equation*}
  \langle \ov{happy}^{(r)}|\ov{kids}\rangle \langle \ov{happy}^{(l)}|\ov{play}^{(l)} \rangle 
  \langle \ov{play}^{(r)}|\ov{games}\rangle \ov{play}^{(m)} 
\end{equation*}

Depending on how one measures the distance between two sentences, this is a very unwelcome effect, to say the least. When using cosine distance, the meaning of all sentences with `play' as the verb will be exactly the same and equal to the middle component of the  `play' tensor. For example, the sentence  ``trembling  shadows play hide-and-seek'' will have the same meaning  as our example sentence. Similarly, the comparison of two arbitrary transitive sentences will be reduced to comparing just the middle components of their verb tensors, completely ignoring any surrounding context. The use of Euclidean distance instead of cosine would slightly improve things, since now we would be at least able to also detect differences in the magnitude between the two middle components. Unfortunately, this metric has been proved not very appropriate for distributional models of meaning, since in the vastness of a highly dimensional space every point ends up to be almost equidistant from all the others. As a result, most implementations of distributional models prefer the more relaxed metric of cosine distance which is length-invariant. Table \ref{tbl:cons} presents the consequences of separability in a number of grammatical constructs.

\renewcommand{\arraystretch}{2.0}
\begin{table}[b]
  \begin{center}
  \footnotesize
  \begin{tabular}{l|l|c}
     \hline
     \textbf{Structure} & \textbf{Simplification} & \textbf{Cos-measured result} \\
     \hline\hline
      adjective-noun & $\overline{adj} \times \ov{noun} = (\ov{adj}^{(l)} \otimes \ov{adj}^{(r)}) \times \ov{noun} = \langle \ov{adj}^{(r)}|\ov{noun}\rangle \cdot \ov{adj}^{(l)}$ & $\ov{adj}^{(l)}$  \\
      \hline
     intrans. sentence & $\ov{subj} \times \overline{verb} = \ov{subj} \times (\ov{verb}^{(l)} \otimes \ov{verb}^{(r)}) = \langle \ov{subj}|\ov{verb}^{(l)}\rangle \cdot \ov{verb}^{(r)}$ & $\ov{verb}^{(r)}$  \\
     \hline
     verb-object & $\overline{verb} \times \ov{obj} = (\ov{verb}^{(l)} \otimes \ov{verb}^{(r)}) \times \ov{obj} = \langle \ov{verb}^{(r)}|\ov{obj}\rangle \cdot \ov{verb}^{(l)}$ &  $\ov{verb}^{(l)}$ \\
     \hline
     transitive sentence & $\begin{array}{r l} \ov{subj} \times \overline{verb} \times \ov{obj} =  
            \ov{subj} \times (\ov{verb}^{(l)} \otimes \ov{verb}^{(m)} \otimes \ov{verb}^{(r)}) \times \ov{obj} & = \\ 
            \langle \ov{subj}|\ov{verb}^{(l)}\rangle \cdot \langle \ov{verb}^{(r)}|\ov{obj}\rangle \cdot \ov{verb}^{(m)} \end{array}$ &  $\ov{verb}^{(m)}$\\
     \hline
  \end{tabular}
  \normalsize
  \end{center}

  \caption{Consequences of separability in various grammatical structures. Superscripts $(l)$, $(m)$ and $(r)$ refer to left-hand, middle, and right-hand component of a separable tensor}
  \label{tbl:cons}
\end{table}
  \renewcommand{\arraystretch}{1.0}
 
\section{Concrete models for verb tensors}
\label{sec:models}

Whereas for the vector representations of atomic words of  language one can use the much-experimented-with methods of distributional semantics, the tensor representations of relational words is a by-product of the categorical framework whose concrete instantiations are still being investigated. A number of  concrete  implementations have been proposed so far, e.g. see \cite{GrefenSadr1,kartsaklis2012,GrefSadrBarIWCS13,KartsaklisEMNLP}. These constructions vary  from corpus-based methods to machine learning techniques. One problem that researchers have had to address is that tensors of order higher than 2 are difficult to create and manipulate. A transitive verb, for example, is represented by a cuboid living in $W^{\otimes 3}$; if the cardinality of our basic vector space is 1000 (and assuming a standard floating-point representation of 8 bytes per real number), the space required for just a single verb becomes 8 gigabytes. A workaround to this issue is to initially create the verb as a matrix, and then expand it to a tensor of higher order by applying Frobenius $\Delta$ operators--that is, leaving one or more dimensions of the resulting tensor empty (filled with zeros). 

A simple and intuitive way to create a matrix for a relational word is to structurally mix the arguments with which this word appears in the training corpus \cite{GrefenSadr1}. For a transtive verb, this would be given us:

\vspace{-0.1cm}
\begin{equation}
  \overline{verb} = \sum_i (\ov{subject_i} \otimes \ov{object_i})
  \label{equ:rel}
\end{equation}
\vspace{-0.2cm}

\noindent where $\ov{subject_i}$ and $\ov{object_i}$ are the vectors of the subject/object pair for the $i$th occurrence of the verb in the corpus. The above technique seems to naturally  result in an entangled matrix, assuming that the family of subject vectors exhibit a sufficient degree of linear independence, and the same is true for the family of object vectors. Compare this to a  straightforward variation which naturally results in a separable matrix, as follows:

\vspace{-0.2cm}
\begin{equation}
  \overline{verb} = \left(\sum_i \ov{subject_i}\right) \otimes \left(\sum_i \ov{object_i}\right)
  \label{equ:sep}
\end{equation}

In what follows, we  present a number of  methods to embed the above verbs from tensors of order 2 to tensors of higher order, as required by the categorical framework. 

%

\paragraph{Relational} In \cite{GrefenSadr1}, the order of a sentence space depends on the arity of the verb of the sentence; for a transitive sentence the result will be a matrix, for an intransitive one it will be a vector, and so on. For the transitive case, the authors expand the original verb matrix to a tensor of order 4 (since now $S=N\otimes N$, the original $N\otimes S \otimes N$ space becomes $N^{\otimes 4}$) by copying both dimensions using Frobenius $\Delta$ operators as shown below: 

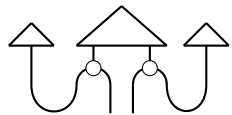
\ctikzfig{rel}

Linear-algebraically, the meaning of a transitive sentence is a matrix living in $W \otimes W$ obtained by the following equation:

\begin{equation}
  \overline{subj~verb~obj} = (subj \otimes obj) \odot \overline{verb}
\end{equation}

\noindent where the symbol $\odot$ denotes element-wise multiplication. 

\paragraph{Frobenius} The above method has the limitation that sentences of different structures live in spaces of different tensor orders, so a direct comparison thereof is not  possible.  As a solution, Kartsaklis et al. \cite{kartsaklis2012} propose the copying of only one dimension of the original matrix, which leads to the following two possibilities:

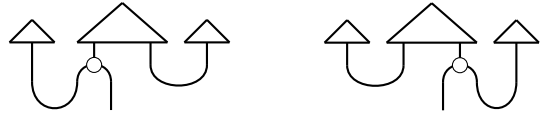
\begin{center}
\begin{tabular}{ccc}
 
\begin{tikzpicture}
	\begin{pgfonlayer}{nodelayer}
		\node [style=none] (0) at (-2, 2.75) {};
		\node [style=none] (1) at (-6, 2) {};
		\node [style=none] (2) at (1.75, 2) {};
		\node [style=none] (3) at (-7.75, 1) {};
		\node [style=none] (4) at (-7, 1) {};
		\node [style=none] (5) at (-6, 1) {};
		\node [style=none] (6) at (-5, 1) {};
		\node [style=none] (7) at (-4, 1) {};
		\node [style=none] (8) at (-3.25, 1) {};
		\node [style=none] (9) at (-0.75, 1) {};
		\node [style=none] (10) at (0, 1) {};
		\node [style=none] (11) at (0.75, 1) {};
		\node [style=none] (12) at (1.75, 1) {};
		\node [style=none] (13) at (2.75, 1) {};
		\node [draw, circle, minimum size=0.2 cm, fill=white, style=none] (14) at (-3.25, -0) {};
		\node [style=none] (15) at (-0.7500001, -0) {};
		\node [style=none] (16) at (1.750001, -0) {};
		\node [style=none] (17) at (-6, -0.7500001) {};
		\node [style=none] (18) at (-4, -0.7500001) {};
		\node [style=none] (19) at (-2.5, -0.7500001) {};
		\node [style=none] (20) at (-2.5, -2) {};
	\end{pgfonlayer}
	\begin{pgfonlayer}{edgelayer}
		\draw [thick, bend left=270, looseness=1.25] (15.center) to (16.center);
		\draw [thick, bend left=90, looseness=1.75] (18.center) to (19.center);
		\draw [thick, looseness=0.00] (8.center) to (14.center);
		\draw [thick, looseness=0.00] (7.center) to (0.center);
		\draw [thick, looseness=0.00] (11.center) to (13.center);
		\draw [thick, looseness=0.00] (11.center) to (2.center);
		\draw [thick, looseness=0.00] (5.center) to (17.center);
		\draw [thick, bend left=270, looseness=2.00] (17.center) to (18.center);
		\draw [thick, looseness=0.00] (4.center) to (1.center);
		\draw [thick, looseness=0.00] (4.center) to (6.center);
		\draw [thick, looseness=0.00] (0.center) to (10.center);
		\draw [thick, looseness=0.00] (7.center) to (10.center);
		\draw [thick, looseness=0.00] (12.center) to (16.center);
		\draw [thick, looseness=0.00] (19.center) to (20.center);
		\draw [thick, looseness=0.00] (9.center) to (15.center);
		\draw [thick, looseness=0.00] (1.center) to (6.center);
		\draw [thick, looseness=0.00] (2.center) to (13.center);
	\end{pgfonlayer}
\end{tikzpicture}}
 &~~~& 
\begin{tikzpicture}
	\begin{pgfonlayer}{nodelayer}
		\node [style=none] (0) at (-2, 2.75) {};
		\node [style=none] (1) at (-5.75, 2) {};
		\node [style=none] (2) at (1.75, 2) {};
		\node [style=none] (3) at (-4, 1) {};
		\node [style=none] (4) at (-3.25, 1) {};
		\node [style=none] (5) at (-0.75, 1) {};
		\node [style=none] (6) at (0, 1) {};
		\node [style=none] (7) at (0.75, 1) {};
		\node [style=none] (8) at (1.75, 1) {};
		\node [style=none] (9) at (2.75, 1) {};
		\node [style=none] (10) at (-6.75, 0.9999998) {};
		\node [style=none] (11) at (-5.75, 0.9999998) {};
		\node [style=none] (12) at (-4.75, 0.9999998) {};
		\node [style=none] (13) at (-5.75, -0) {};
		\node [style=none] (14) at (-3.25, -0) {};
		\node [draw, circle, minimum size=0.2 cm, fill=white, style=none] (15) at (-0.7500001, -0) {};
		\node [style=none] (16) at (-1.5, -0.7500001) {};
		\node [style=none] (17) at (0, -0.7500001) {};
		\node [style=none] (18) at (1.750001, -0.7500001) {};
		\node [style=none] (19) at (-1.5, -2) {};
	\end{pgfonlayer}
	\begin{pgfonlayer}{edgelayer}
		\draw [thick, looseness=0.00] (16.center) to (19.center);
		\draw [thick, looseness=0.00] (3.center) to (6.center);
		\draw [thick, looseness=0.00] (7.center) to (9.center);
		\draw [thick, looseness=0.00] (10.center) to (1.center);
		\draw [thick, bend left=270, looseness=2.25] (17.center) to (18.center);
		\draw [thick, bend left=90, looseness=1.75] (16.center) to (17.center);
		\draw [thick, looseness=0.00] (7.center) to (2.center);
		\draw [thick, looseness=0.00] (3.center) to (0.center);
		\draw [thick, looseness=0.00] (5.center) to (15.center);
		\draw [thick, looseness=0.00] (4.center) to (14.center);
		\draw [thick, looseness=0.00] (2.center) to (9.center);
		\draw [thick, looseness=0.00] (10.center) to (12.center);
		\draw [thick, looseness=0.00] (1.center) to (12.center);
		\draw [thick, looseness=0.00] (8.center) to (18.center);
		\draw [thick, looseness=0.00] (0.center) to (6.center);
		\draw [thick, bend right=90, looseness=1.25] (13.center) to (14.center);
		\draw [thick, looseness=0.00] (11.center) to (13.center);
	\end{pgfonlayer}
\end{tikzpicture}}

\end{tabular}
\end{center}

\noindent
The result is now a vector, computed in the following way, respectively for each case:

\vspace{-0.5cm}
\begin{eqnarray}
  \text{\bf Copy-subject:} \qquad \ov{subj~verb~obj} = \ov{subj} \odot (\overline{verb} \times \ov{obj})\label{equ:copysbj} \\
   \text{\bf Copy-object:}\qquad  \ov{subj~verb~obj} = \ov{obj} \odot (\overline{verb}^\mathsf{T} \times \ov{subj})\label{equ:copyobj}
\end{eqnarray}

Each one of the vectors obtained from Eqs. \ref{equ:copysbj} and \ref{equ:copyobj} above addresses a partial interaction of the verb with each argument. It is reasonable then to further combine them in order to get a more complete representation of the verb meaning (and hence the sentence meaning). We therefore define three more models, in which this combination is achieved through vector addition ({\bf Frobenius additive}), element-wise multiplication ({\bf Frobenius multiplicative}), and tensor product ({\bf Frobenius tensored}) of the above. 

We conclude this section with two important comments. First, although the use of a matrix for representing a transitive verb might originally seem as a violation of the functorial relation with a pregroup grammar, this is not the case in practice; the functorial relation is restored through the use of the Frobenius operators, which produce a tensor of the correct order, as required by the grammatical type. Furthermore, this notion of ``inflation'' has the additional advantage that can also work from a reversed perspective: a matrix created by Eq. \ref{equ:rel} can be seen as an order-3 tensor originally in $N\otimes S \otimes N$ where the $S$ dimension has been discarded by a $\zeta$ Frobenius map. Using this approach, Sadrzadeh and colleagues provide intuitive analyses for wh-movement phenomena and discuss compositional treatments of constructions containing relative pronouns \cite{sadrzadeh2013frobenius, sadrzadeh2014frobenius}.

Finally, we would like to stress out the fact that, despite of the actual level of entanglement in our original verb matrix created by Eq. \ref{equ:rel}, the use of Frobenius operators as described above equips the inflated verb tensors with an extra level of entanglement in any case. As we will see in Sect. \ref{sec:discussion} when discussing the results of the experimental work, this detail will be proven very important in practice.

\section{Experiments}
\label{sec:exp}

\subsection{Creating a semantic space}
\label{sec:space}

Our basic vector space is trained from the ukWaC corpus \cite{ukwac}, originally using as a basis the 2,000 content words with the highest frequency (but excluding a list of stop words as well as the 50 most frequent content words since they exhibit low information content). As context we considered a 5-word window from either side of the target word, whereas for our weighting scheme we used local mutual information (i.e. point-wise mutual information multiplied by raw counts). The vector space was normalized and projected onto a 300-dimensional space using singular value decomposition (SVD). These choices are based on our best results in a number of previous experiments \cite{KartsaklisEMNLP, KartsaklisACL}. 

\subsection{Detecting sentence similarity}

In this section we test the various compositional models of Sect. \ref{sec:models} in two similarity tasks involving pairs of transitive sentences; for each pair, we construct composite vectors for the two sentences, and then we measure their semantic similarity using cosine distance and Euclidean distance. We then evaluate the correlation of each model's performance with human judgements, using Spearman's $\rho$. In the first task \cite{GrefenSadr1}, the sentences to be compared are constructed using the same subject and object and semantically correlated verbs, such as `spell' and `write'; for example, `pupils write letters' is compared with `pupils spell letters'. The dataset consists of 200 sentence pairs. 

We are especially interested in measuring the level of entanglement in our verb matrices as these are created by Eq. \ref{equ:rel}. In order to achieve that, we compute the \textit{rank-1 approximation} of all verbs in our dataset. Given a verb matrix $\overline{verb}$, we first compute its SVD so that $\overline{verb} = \textbf{U} \boldsymbol{\Sigma} \textbf{V}\mathsf{^T}$, and then we approximate this matrix by using only the highest eigenvalue and the related left and right singular vectors, so that $\overline{verb}_{R1} = \textbf{U}_1 \boldsymbol{\Sigma}_1 \textbf{V}^\mathsf{T}_1$.
%
%
We compare the composite vectors created by the original matrix (Eq. \ref{equ:rel}), their rank-1 approximations, and the results of the separable model of Eq. \ref{equ:sep}. We also use a number of baselines: in the `verbs-only' model,  we compare only the verbs (without composing them with the context), while in  the additive and multiplicative models we construct the sentence vectors by simply adding and element-wise multiplying the distributional vectors of their words. 

The results (Table \ref{tbl:exp1}) revealed a striking similarity in the performances of the  entangled  and  separable versions. Using cosine distance, all three models (relational,  rank-1 approximation,  separable model) have essentially the same behaviour; with Euclidean distance, the relational model performs again the same as its rank-1 approximation, while this time the separable model is lower. 

\begin{table}[h!]
  \centering
  \small
  \begin{tabular}{l|cc}
    \hline
    \textbf{Model} & \textbf{$\rho$ with cos} & \textbf{$\rho$ with Eucl.} \\
    \hline\hline
    Verbs only          & 0.329 & 0.138 \\
    Additive            & 0.234 & 0.142 \\
    Multiplicative      & 0.095 & 0.024 \\
    \hline
    Relational          & 0.400 & 0.149 \\
    Rank-1 approx. of relational  & 0.402 & 0.149 \\
    Separable                & 0.401 & 0.090 \\
    \hline
    Copy-subject             & 0.379 & 0.115\\
    Copy-object              & 0.381 & 0.094\\
    Frobenius additive       & \textbf{0.405} & 0.125\\
    Frobenius multiplicative & 0.338 & 0.034\\
    Frobenius tensored       & \textbf{0.415} & 0.010\\
    \hline
    Human agreement       & \multicolumn{2}{c}{0.60} \\    
    \hline
  \end{tabular}
  \normalsize
  \caption{Results for the first dataset (same subjects/objects, semantically related verbs)}.
  \label{tbl:exp1}
\end{table}

The inevitable conclusion that Eq. \ref{equ:rel} actually produces a separable matrix was further confirmed by an additional experiment: we calculated the average cosine similarity of the original matrices with their rank-1 approximations, a computation that revealed a similarity as high as 0.99. Since this result might obviously depend on the form of the noun vectors used for creating the verb matrix, this last experiment was repeated with a number of variations of our basic vector space, getting in every case similarities between verb matrices and their rank-1 approximations higher than 0.97. The observed behaviour can only be explained with the presence of a very high level of linear dependence between the subject vectors and between the object vectors. If every subject vector can be expressed as a linear combination of a small number of other vectors (and the same is true for the family of object vectors), then this would drastically reduce the entanglement of the matrix to the level that it is in effect separable. 

Our observations are also confirmed in the second sentence similarity task. Here, we use a variation of one of the datasets in \cite{KartsaklisEMNLP}, consisting of 108 pairs of transitive sentences. The difference with our first task is that now the sentences of a pair are unrelated in a word level, i.e. subjects, objects, and verbs are all different. The results for this second experiment are presented in Table \ref{tbl:exp2}. 

\begin{table}[h]
  \centering
  \small
  \begin{tabular}{l|cc}
    \hline
    \textbf{Model} & \textbf{$\rho$ with cos} & \textbf{$\rho$ with Eucl.} \\
    \hline\hline
    Verbs only           & 0.449      & 0.392\\
    Additive             & 0.581      & 0.542\\
    Multiplicative       & 0.287      & 0.109\\
    \hline
    Relational           & 0.334      & 0.173 \\
    Rank-1 approx. of relational & 0.333 & 0.175 \\
    Separable            & 0.332      & 0.105\\
    \hline
    Copy-subject          & 0.427     & 0.096\\
    Copy-object           & 0.198     & 0.144\\
    Frobenius additive    & 0.428    & 0.117\\
    Frobenius multiplicative &  0.302 & 0.041 \\
    Frobenius tensored    &  0.332   & 0.042\\
    \hline
    Human agreement       & \multicolumn{2}{c}{0.66} \\
    \hline
  \end{tabular}
  \normalsize
  \caption{Results for the second dataset (different subjects, objects and verbs)}.
  \label{tbl:exp2}
\end{table}

As a general observation about the performance of the various models in the two tasks, we note the high scores achieved by the Frobenius models when one uses the preferred method of measurement, that of cosine similarity. Especially the \textbf{Frobenius additive} has been proved to perform better than the Relational model, having the additional advantage that it allows comparison between sentences of different structures (since every sentence vector lives in $W$).

\vspace{-0.1cm}
\section{Discussion}
\label{sec:discussion}

The experiments of Sect. \ref{sec:exp} revealed an unwelcome property of a method our colleagues and we have used in the past for creating verb tensors in the context of compositional models \cite{GrefenSadr1,kartsaklis2012,KartsaklisEMNLP}. The fact that the verb matrix is in effect separable introduces a number of simplifications in the models presented in Sect. \ref{sec:models}. More specifically, the Relational model of \cite{GrefenSadr1} is reduced to the following:

\vspace{0.3cm}
\begin{tabular}{cc}

\begin{tikzpicture}
	\begin{pgfonlayer}{nodelayer}
		\node [style=none] (0) at (-8.25, 2) {};
		\node [style=none] (1) at (-5.5, 2) {};
		\node [style=none] (2) at (-3, 2) {};
		\node [style=none] (3) at (-0.5, 2) {};
		\node [style=none] (4) at (3, 2) {};
		\node [style=none] (5) at (6, 2) {};
		\node [style=none] (6) at (8.5, 2) {};
		\node [style=none] (7) at (11.5, 2) {};
		\node [style=none] (8) at (-9.25, 1) {};
		\node [style=none] (9) at (-8.25, 1) {};
		\node [style=none] (10) at (-7.25, 1) {};
		\node [style=none] (11) at (-6.5, 1) {};
		\node [style=none] (12) at (-5.5, 1) {};
		\node [style=none] (13) at (-4.5, 1) {};
		\node [style=none] (14) at (-4, 1) {};
		\node [style=none] (15) at (-3, 1) {};
		\node [style=none] (16) at (-2, 1) {};
		\node [style=none] (17) at (-1.5, 1) {};
		\node [style=none] (18) at (-0.5, 1) {};
		\node [style=none] (19) at (0.5, 1) {};
		\node [style=none] (20) at (2, 1) {};
		\node [style=none] (21) at (3, 1) {};
		\node [style=none] (22) at (4, 1) {};
		\node [style=none] (23) at (5, 1) {};
		\node [style=none] (24) at (6, 1) {};
		\node [style=none] (25) at (7, 1) {};
		\node [style=none] (26) at (7.5, 1) {};
		\node [style=none] (27) at (8.5, 1) {};
		\node [style=none] (28) at (9.5, 1) {};
		\node [style=none] (29) at (10.5, 1) {};
		\node [style=none] (30) at (11.5, 1) {};
		\node [style=none] (31) at (12.5, 1) {};
		\node [draw, circle, minimum size=0.2 cm, fill=white, style=none] (32) at (-5.5, -0) {};
		\node [draw, circle, minimum size=0.2 cm, fill=white, style=none] (33) at (-3, -0) {};
		\node [style=none] (34) at (1.25, -0) {$=$};
		\node [draw, circle, minimum size=0.2 cm, fill=white, style=none] (35) at (4.5, -0.32) {};
		\node [draw, circle, minimum size=0.2 cm, fill=white, style=none] (36) at (10, -0.32) {};
		\node [style=none] (37) at (-8.25, -0.75) {};
		\node [style=none] (38) at (-6.25, -0.75) {};
		\node [style=none] (39) at (-4.75, -0.75) {};
		\node [style=none] (40) at (-3.75, -0.75) {};
		\node [style=none] (41) at (-2.25, -0.75) {};
		\node [style=none] (42) at (-0.5, -0.75) {};
		\node [style=none] (43) at (-4.75, -2) {};
		\node [style=none] (44) at (-3.75, -2) {};
		\node [style=none] (45) at (4.5, -2) {};
		\node [style=none] (46) at (10, -2) {};
	\end{pgfonlayer}
	\begin{pgfonlayer}{edgelayer}
		\draw [thick, looseness=0.00] (20.center) to (22.center);
		\draw [thick, bend left=90, looseness=1.75] (38.center) to (39.center);
		\draw [thick, looseness=0.00] (7.center) to (31.center);
		\draw [thick, looseness=0.00] (11.center) to (13.center);
		\draw [thick] (36.center) to (46.center);
		\draw [thick, looseness=0.00] (29.center) to (31.center);
		\draw [thick, looseness=0.00] (5.center) to (25.center);
		\draw [thick, looseness=0.00] (1.center) to (13.center);
		\draw [thick, bend left=270, looseness=2.25] (41.center) to (42.center);
		\draw [thick, looseness=0.00] (18.center) to (42.center);
		\draw [thick, looseness=0.00] (8.center) to (0.center);
		\draw [thick, looseness=0.00] (9.center) to (37.center);
		\draw [thick, looseness=0.00] (2.center) to (16.center);
		\draw [thick, looseness=0.00] (26.center) to (28.center);
		\draw [thick, looseness=0.00] (17.center) to (3.center);
		\draw [thick, looseness=0.00] (17.center) to (19.center);
		\draw [thick, looseness=0.00] (6.center) to (28.center);
		\draw [thick, bend left=270, looseness=1.50] (21.center) to (24.center);
		\draw [thick, looseness=0.00] (23.center) to (25.center);
		\draw [thick, looseness=0.00] (14.center) to (2.center);
		\draw [thick, looseness=0.00] (14.center) to (16.center);
		\draw [thick, looseness=0.00] (0.center) to (10.center);
		\draw [thick, looseness=0.00] (3.center) to (19.center);
		\draw [thick, looseness=0.00] (39.center) to (43.center);
		\draw [thick, bend left=90, looseness=1.75] (40.center) to (41.center);
		\draw [thick, looseness=0.00] (20.center) to (4.center);
		\draw [thick, looseness=0.00] (11.center) to (1.center);
		\draw [thick, looseness=0.00] (23.center) to (5.center);
		\draw [thick, looseness=0.00] (29.center) to (7.center);
		\draw [thick, bend right=90, looseness=1.50] (27.center) to (30.center);
		\draw [thick, looseness=0.00] (26.center) to (6.center);
		\draw [thick, looseness=0.00] (12.center) to (32.center);
		\draw [thick, bend left=270, looseness=2.00] (37.center) to (38.center);
		\draw [thick, looseness=0.00] (4.center) to (22.center);
		\draw [thick, looseness=0.00] (40.center) to (44.center);
		\draw [thick, looseness=0.00] (8.center) to (10.center);
		\draw [thick, looseness=0.00] (15.center) to (33.center);
		\draw [thick] (35.center) to (45.center);
	\end{pgfonlayer}
\end{tikzpicture}}
 &
  $\overline{subj~verb~obj} = (\ov{subj} \odot \ov{verb}^{(l)}) \otimes (\ov{verb}^{(r)} \odot \ov{obj})$\\
\end{tabular}
\vspace{0.3cm}

\noindent
Furthermore, the Frobenius models of \cite{kartsaklis2012} get these forms:

\begin{center}
\begin{tabular}{ccc}
 
\begin{tikzpicture}
	\begin{pgfonlayer}{nodelayer}
		\node [style=none] (0) at (-6, 2) {};
		\node [style=none] (1) at (-3.25, 2) {};
		\node [style=none] (2) at (-0.75, 2) {};
		\node [style=none] (3) at (1.75, 2) {};
		\node [style=none] (4) at (5, 2) {};
		\node [style=none] (5) at (8, 2) {};
		\node [style=none] (6) at (-7, 1) {};
		\node [style=none] (7) at (-6, 1) {};
		\node [style=none] (8) at (-5, 1) {};
		\node [style=none] (9) at (-4.25, 1) {};
		\node [style=none] (10) at (-3.25, 1) {};
		\node [style=none] (11) at (-2.25, 1) {};
		\node [style=none] (12) at (-1.75, 1) {};
		\node [style=none] (13) at (-0.75, 1) {};
		\node [style=none] (14) at (0.25, 1) {};
		\node [style=none] (15) at (0.75, 1) {};
		\node [style=none] (16) at (1.75, 1) {};
		\node [style=none] (17) at (2.75, 1) {};
		\node [style=none] (18) at (4, 1) {};
		\node [style=none] (19) at (5, 1) {};
		\node [style=none] (20) at (6, 1) {};
		\node [style=none] (21) at (7, 1) {};
		\node [style=none] (22) at (8, 1) {};
		\node [style=none] (23) at (9, 1) {};
		\node [draw, circle, minimum size=0.2 cm, fill=white, style=none] (24) at (-3.25, -0) {};
		\node [style=none] (25) at (-0.7500001, -0) {};
		\node [style=none] (26) at (1.750001, -0) {};
		\node [style=none] (27) at (3.5, -0) {$=$};
		\node [draw, circle, minimum size=0.2 cm, fill=white, style=none] (28) at (6.5, -0.32) {};
		\node [style=none] (29) at (-6, -0.7500001) {};
		\node [style=none] (30) at (-4, -0.7500001) {};
		\node [style=none] (31) at (-2.5, -0.7500001) {};
		\node [style=none] (32) at (-2.5, -2) {};
		\node [style=none] (33) at (6.5, -2) {};
	\end{pgfonlayer}
	\begin{pgfonlayer}{edgelayer}
		\draw [thick] (28.center) to (33.center);
		\draw [thick, bend left=270, looseness=1.50] (19.center) to (22.center);
		\draw [thick, looseness=0.00] (10.center) to (24.center);
		\draw [thick, looseness=0.00] (21.center) to (5.center);
		\draw [thick, bend left=270, looseness=2.00] (29.center) to (30.center);
		\draw [thick, looseness=0.00] (7.center) to (29.center);
		\draw [thick, looseness=0.00] (18.center) to (20.center);
		\draw [thick, looseness=0.00] (18.center) to (4.center);
		\draw [thick, looseness=0.00] (9.center) to (11.center);
		\draw [thick, looseness=0.00] (21.center) to (23.center);
		\draw [thick, looseness=0.00] (12.center) to (14.center);
		\draw [thick, looseness=0.00] (12.center) to (2.center);
		\draw [thick, looseness=0.00] (6.center) to (0.center);
		\draw [thick, looseness=0.00] (3.center) to (17.center);
		\draw [thick, looseness=0.00] (6.center) to (8.center);
		\draw [thick, looseness=0.00] (5.center) to (23.center);
		\draw [thick, looseness=0.00] (16.center) to (26.center);
		\draw [thick, looseness=0.00] (31.center) to (32.center);
		\draw [thick, looseness=0.00] (15.center) to (17.center);
		\draw [thick, looseness=0.00] (1.center) to (11.center);
		\draw [thick, looseness=0.00] (9.center) to (1.center);
		\draw [thick, looseness=0.00] (2.center) to (14.center);
		\draw [thick, looseness=0.00] (4.center) to (20.center);
		\draw [thick, bend left=270, looseness=1.25] (25.center) to (26.center);
		\draw [thick, bend left=90, looseness=1.75] (30.center) to (31.center);
		\draw [thick, looseness=0.00] (0.center) to (8.center);
		\draw [thick, looseness=0.00] (13.center) to (25.center);
		\draw [thick, looseness=0.00] (15.center) to (3.center);
	\end{pgfonlayer}
\end{tikzpicture}}
 & & 
\begin{tikzpicture}
	\begin{pgfonlayer}{nodelayer}
		\node [style=none] (0) at (-6, 2) {};
		\node [style=none] (1) at (-3.5, 2) {};
		\node [style=none] (2) at (-1, 2) {};
		\node [style=none] (3) at (1.75, 2) {};
		\node [style=none] (4) at (5, 2) {};
		\node [style=none] (5) at (8, 2) {};
		\node [style=none] (6) at (-7, 1) {};
		\node [style=none] (7) at (-6, 1) {};
		\node [style=none] (8) at (-5, 1) {};
		\node [style=none] (9) at (-4.5, 1) {};
		\node [style=none] (10) at (-3.5, 1) {};
		\node [style=none] (11) at (-2.5, 1) {};
		\node [style=none] (12) at (-2, 1) {};
		\node [style=none] (13) at (-1, 1) {};
		\node [style=none] (14) at (0, 1) {};
		\node [style=none] (15) at (0.75, 1) {};
		\node [style=none] (16) at (1.75, 1) {};
		\node [style=none] (17) at (2.75, 1) {};
		\node [style=none] (18) at (4, 1) {};
		\node [style=none] (19) at (5, 1) {};
		\node [style=none] (20) at (6, 1) {};
		\node [style=none] (21) at (7, 1) {};
		\node [style=none] (22) at (8, 1) {};
		\node [style=none] (23) at (9, 1) {};
		\node [style=none] (24) at (-6.000001, -0) {};
		\node [style=none] (25) at (-3.5, -0) {};
		\node [draw, circle, minimum size=0.2 cm, fill=white, style=none] (26) at (-1, -0) {};
		\node [style=none] (27) at (3.5, -0) {$=$};
		\node [draw, circle, minimum size=0.2 cm, fill=white, style=none] (28) at (6.5, -0.32) {};
		\node [style=none] (29) at (-1.75, -0.7500001) {};
		\node [style=none] (30) at (-0.25, -0.7500001) {};
		\node [style=none] (31) at (1.75, -0.7500001) {};
		\node [style=none] (32) at (-1.75, -2) {};
		\node [style=none] (33) at (6.5, -2) {};
	\end{pgfonlayer}
	\begin{pgfonlayer}{edgelayer}
		\draw [thick] (28.center) to (33.center);
		\draw [thick, bend left=270, looseness=1.50] (19.center) to (22.center);
		\draw [thick, looseness=0.00] (13.center) to (26.center);
		\draw [thick, looseness=0.00] (21.center) to (5.center);
		\draw [thick, bend right=270, looseness=2.00] (31.center) to (30.center);
		\draw [thick, looseness=0.00] (16.center) to (31.center);
		\draw [thick, looseness=0.00] (18.center) to (20.center);
		\draw [thick, looseness=0.00] (18.center) to (4.center);
		\draw [thick, looseness=0.00] (14.center) to (12.center);
		\draw [thick, looseness=0.00] (21.center) to (23.center);
		\draw [thick, looseness=0.00] (11.center) to (9.center);
		\draw [thick, looseness=0.00] (11.center) to (1.center);
		\draw [thick, looseness=0.00] (17.center) to (3.center);
		\draw [thick, looseness=0.00] (0.center) to (6.center);
		\draw [thick, looseness=0.00] (17.center) to (15.center);
		\draw [thick, looseness=0.00] (5.center) to (23.center);
		\draw [thick, looseness=0.00] (7.center) to (24.center);
		\draw [thick, looseness=0.00] (29.center) to (32.center);
		\draw [thick, looseness=0.00] (8.center) to (6.center);
		\draw [thick, looseness=0.00] (2.center) to (12.center);
		\draw [thick, looseness=0.00] (14.center) to (2.center);
		\draw [thick, looseness=0.00] (1.center) to (9.center);
		\draw [thick, looseness=0.00] (4.center) to (20.center);
		\draw [thick, bend right=270, looseness=1.25] (25.center) to (24.center);
		\draw [thick, bend right=90, looseness=1.75] (30.center) to (29.center);
		\draw [thick, looseness=0.00] (3.center) to (15.center);
		\draw [thick, looseness=0.00] (10.center) to (25.center);
		\draw [thick, looseness=0.00] (8.center) to (0.center);
	\end{pgfonlayer}
\end{tikzpicture}}

\end{tabular}
\end{center}

\noindent
which means, for example, that the actual equation behind the successful Frobenius additive model is 

\begin{equation}
\ov{subj~verb~obj} = (\ov{subj} \odot \ov{verb}^{(l)}) + (\ov{verb}^{(r)} \odot \ov{obj})
  \label{equ:frobadd}
\end{equation}

Despite the simplifications presented above, note that none of these models degenerates to the level of producing ``constant'' vectors or matrices, as argued for in Sect. \ref{sec:cons}. Indeed, especially in the first task (Table \ref{tbl:exp1}) the Frobenius models present top performance, and the relational models follow closely. The reason behind this lies in the use of Frobenius $\Delta$ operators for copying the original dimensions of the verb matrix, a computation that equipped the fragmented system with flow, although not in the originally intended sense. The compositional structure is still fragmented into two parts, but at least now the copied dimensions provide a means to deliver the results of the two individual computations that take place, one for the left-hand part of the sentence and one for the right-hand part. Let us  see what happens when we use cosine distance in order to compare the matrices of two transitive sentences created with the {\bf Relational} model (the separable version of a verb matrix $\overline{verb}$ is denoted  by $\ov{verb}^{(l)} \otimes \ov{verb}^{(r)}$):

\begin{eqnarray*}
  \left\langle \overline{subj_1~verb_1~obj_1} \middle| \overline{subj_2~verb_2~obj_2} \right\rangle & = \\
   \left\langle ( \ov{subj}_1 \odot \ov{verb}_1^{(l)}) \otimes (\ov{verb}_1^{(r)} \odot \ov{obj}_1 ) \middle|
   ( \ov{subj}_2 \odot \ov{verb}_2^{(l)}) \otimes (\ov{verb}_2^{(r)} \odot \ov{obj}_2 ) \right\rangle & = \\
   \left\langle \ov{subj}_1 \odot \ov{verb}_1^{(l)} \middle| \ov{subj}_2 \odot \ov{verb}_2^{(l)} \right\rangle  \left\langle \ov{verb}_1^{(r)} \odot \ov{obj}_1 \middle| \ov{verb}_2^{(r)} \odot \ov{obj}_2 \right\rangle
\end{eqnarray*}

As also computed and pointed out in \cite{GrefSadrCL}, the two sentences are broken up to a left-hand part and a right-hand part, and two distinct comparisons take place. As long as we compare sentences of the same structure, as we did here, this method is viable. On the other hand, the \textbf{Frobenius} models and their simplifications such as the one in (\ref{equ:frobadd}) do not have this restriction; in principle, all sentences are represented by vectors living in the same space, so any kind of comparison is possible. In case, however, we do compare sentences of the same structure, these models have the additional advantage that also allow comparisons between \textit{different} sentence parts; this can be seen in the dot product of two sentences created by Eq. \ref{equ:frobadd}, which gets the following form:

\begin{eqnarray*}
   \left\langle \ov{subj}_1 \odot \ov{verb}_1^{(l)} \middle| \ov{subj}_2 \odot \ov{verb}_2^{(l)} \right\rangle  +
   \left\langle \ov{subj}_1 \odot \ov{verb}_1^{(l)} \middle| \ov{verb}_2^{(r)} \odot \ov{obj}_2 \right\rangle  & + \\  
   \left\langle \ov{verb}_1^{(r)} \odot \ov{obj}_1 \middle| \ov{subj}_2 \odot \ov{verb}_2^{(l)} \right\rangle +
   \left\langle \ov{verb}_1^{(r)} \odot \ov{obj}_1 \middle| \ov{verb}_2^{(r)} \odot \ov{obj}_2 \right\rangle
\end{eqnarray*}


\section{Using linear regression for entanglement}
\label{sec:lr}

Corpus-based methods for creating tensors of relational words, such as the models presented so far in this paper, are intuitive and easy to implement. As our experimental work shows, however, this convenience comes with a price. In practice, one would expect that more robust machine learning techniques would produce more reliable tensor representations for composition.

In this section we apply linear regression (following \cite{Baroni}) in order to train verb matrices for a variation of our second experiment, in which we compare elementary verb phrases of the form \textit{verb-object} \cite{lapata2010} (so the subjects are dropped). In order to create a matrix for, say, the verb `play', we first collect all instances of the verb occurring with some object in the training corpus, and then we create non-compositional holistic vectors for these elementary verb phrases following exactly the same methodology as if they were words. We now have a dataset with instances of the form $\langle \ov{obj_i}, \ov{play~obj_i} \rangle$ (e.g. the vector of `flute' paired with the holistic vector of `play flute', and so on), that can be used to train a linear regression model in order to produce an appropriate matrix for verb `play'. The premise of a model like this is that the multiplication of the verb matrix with the vector of a new object will produce a result that approximates the distributional behaviour of all these elementary two-word exemplars used in training. For a given verb, this is achieved by using \textit{gradient descent} in order to minimize the total error between the observed vectors and the vectors predicted by the model, expressed by the following quantity:

\begin{equation}
  \frac{1}{2m}\left( \sum_i \overline{verb} \times \ov{object}_i - \ov{verb~object_i} \right)^2
\end{equation}
%

\noindent
where $m$ is the number of training instances. The average cosine similarity between the matrices we got from this method and their rank-1 approximation was only 0.48, showing that in general the level of entanglement produced by this method is reasonably high. This is also confirmed by the results in Table \ref{tbl:exp3}; the rank-1 approximation model presents the worst performance, since, as you might recall from the discussion in Sect. \ref{sec:cons}, separability here means that every verb-object composition is reduced to the left component of the verb matrix, completely ignoring the interaction with the object. 

\begin{table}[h!]
  \centering
  \small
  \begin{tabular}{l|cc}
    \hline
    \textbf{Model} & \textbf{$\rho$ with cos} & \textbf{$\rho$ with Eucl.} \\
    \hline\hline
    Verbs only          & 0.331 & 0.267 \\
    Holistic verb-phrase vectors & 0.403 & 0.214 \\
    Additive            & 0.379 & 0.385 \\
    Multiplicative      & 0.301 & 0.229 \\
    \hline
    Linear regression   & 0.349 & 0.144 \\
    Rank-1 approximation of LR matrices & 0.119 & 0.082 \\
    \hline
    Human agreement       & \multicolumn{2}{c}{0.55} \\
    \hline
  \end{tabular}
  \normalsize
  \caption{Results for the verb-phrase similarity task}
  \label{tbl:exp3}
\end{table}

\vspace{-0.2cm}
\section{Conclusion}
\vspace{-0.2cm}

The current study takes a closer look to an aspect of tensor-based compositional models of meaning that so far had escaped the attention of researchers. The discovery that a number of concrete instantiations of the categorical framework proposed in \cite{coeckeetal} produce relational tensors that are in effect separable stresses the necessity of similar tests for any linear model that follows the same philosophy. Another contribution of this work was that it showed this is not necessarily a bad thing. The involvement of Frobenius operators in the creation of verb tensors equips the compositional structure with the necessary flow, so that a comparison between two sentence vectors can be still carried out between individual parts of each sentence. Therefore, approaches such as the Frobenius additive model proposed in this paper can be still considered as viable and ``easy'' alternatives to more robust machine learning techniques, such as the gradient optimization technique discussed in Sect. \ref{sec:lr}. 

\vspace{-0.3cm} 
\bibliographystyle{eptcs}
\bibliography{refs}

\end{document}